\documentclass[10pt,twocolumn,letterpaper]{article}

\usepackage{cvpr}      
\usepackage{bm}
\usepackage{etoolbox}
\usepackage{silence}
\makeatletter
\robustify\@latex@warning@no@line
\makeatother
\usepackage{authblk}

\definecolor{cvprblue}{rgb}{0.21,0.49,0.74}
\usepackage[pagebackref,breaklinks,colorlinks,allcolors=cvprblue]{hyperref}
\usepackage{endnotes}
\usepackage{multirow}

\title{Instruction-Driven Fusion of Infrared-Visible Images: Tailoring for Diverse Downstream Tasks}

\author[1]{Zengyi Yang}
\author[1]{Yafei Zhang}
\author[1]{Huafeng Li}
\author[2]{Yu Liu}
\affil[1]{Kunming University of Science and Technology}
\affil[2]{Hefei University of Technology}
\affil[ ]{\tt\small zengyiyang0211@gmail.com \hspace{0.2cm} zyfeimail@163.com \hspace{0.2cm} lhfchina99@kust.edu.cn \hspace{0.2cm} yuliu@hfut.edu.cn}

\begin{document}
\maketitle
\begin{abstract}
The primary value of infrared and visible image fusion technology lies in applying the fusion results to downstream tasks. However, existing methods face challenges such as increased training complexity and significantly compromised performance of individual tasks when addressing multiple downstream tasks simultaneously. To tackle this, we propose Task-Oriented Adaptive Regulation (T-OAR), an adaptive mechanism specifically designed for multi-task environments. Additionally, we introduce the Task-related Dynamic Prompt Injection (T-DPI) module, which generates task-specific dynamic prompts from user-input text instructions and integrates them into target representations. This guides the feature extraction module to produce representations that are more closely aligned with the specific requirements of downstream tasks. By incorporating the T-DPI module into the T-OAR framework, our approach generates fusion images tailored to task-specific requirements without the need for separate training or task-specific weights. This not only reduces computational costs but also enhances adaptability and performance across multiple tasks. Experimental results show that our method excels in object detection, semantic segmentation, and salient object detection, demonstrating its strong adaptability, flexibility, and task specificity. This provides an efficient solution for image fusion in multi-task environments, highlighting the technology's potential across diverse applications. \vspace{-4mm}

\end{abstract}    
\section{Introduction}
\label{sec:intro}
\begin{figure}[t]  
	\centering
	\includegraphics[width=0.88\linewidth]{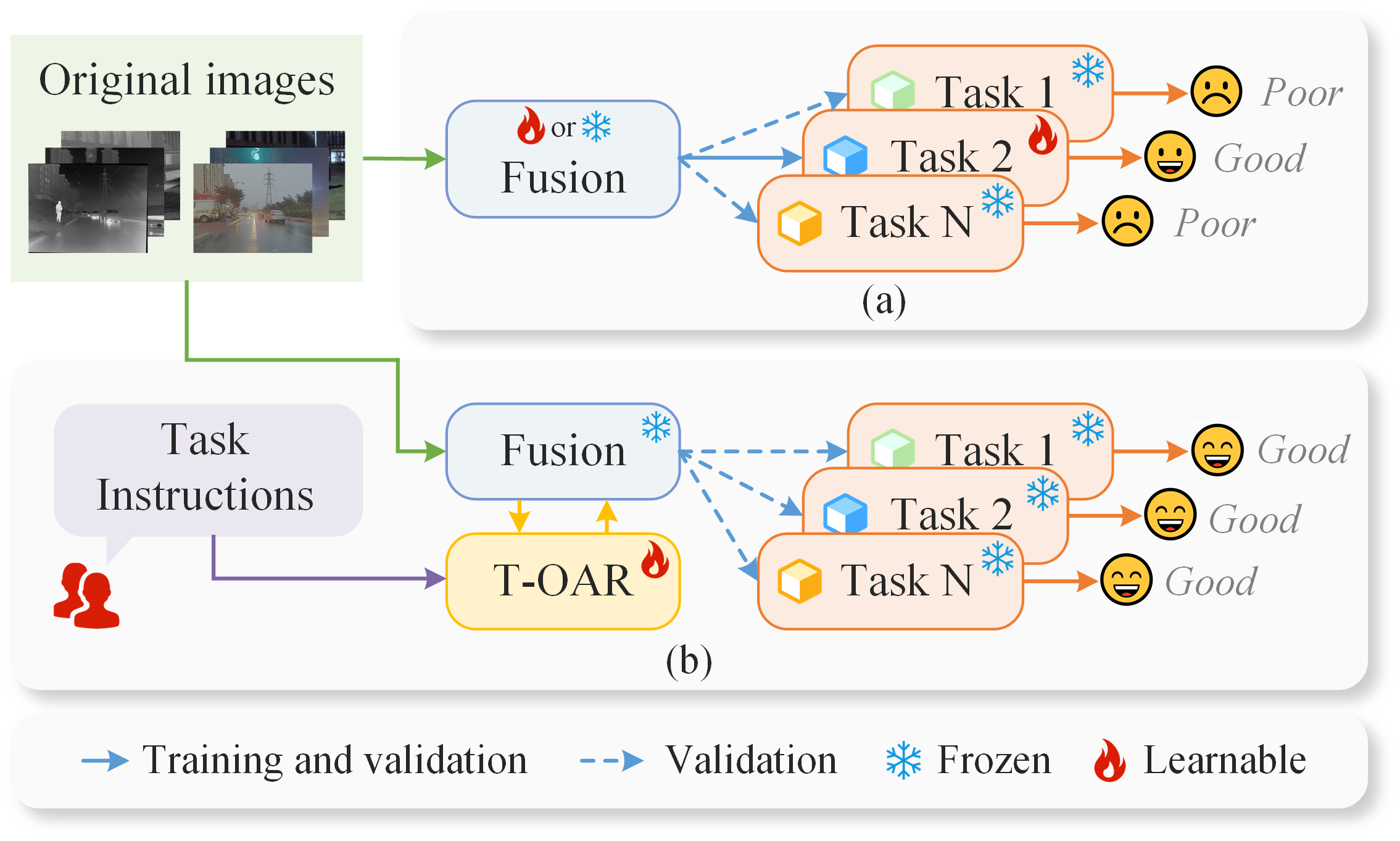}\vspace{-2mm}
	\caption{Comparison of the proposed method with existing approaches. Existing methods (a) perform well only on the specific downstream tasks for which they are trained, thereby limiting their multi-task adaptability. In contrast, our method (b) uses T-OAR with task instructions to fine-tune the fusion network, thereby simultaneously meeting the requirements of multiple downstream tasks without retraining.}\vspace{-4mm}
	\label{fig1}
\end{figure}

Infrared images capture thermal radiation from objects, making them suitable for target detection and recognition in nighttime or adverse weather conditions such as haze, rain, and snow. However, they generally have lower resolution and lack color. Visible images, on the other hand, provide rich color, texture, and detail but suffer from quality loss in low-light or adverse weather conditions. Infrared-visible (IR-VIS) image fusion effectively combines the strengths of both, generating a fused image that retains thermal information from infrared images while incorporating the color and texture details of visible images. This provides a more comprehensive and accurate scene representation. Consequently, IR-VIS image fusion has broad applications in military, aerospace, environmental monitoring, medicine, and other fields, leading to the development of various fusion methods \cite{30,31,32,33,34,35,36,37,40}.

For IR-VIS image fusion, the key value of this technology is its application to downstream visual tasks. As a result, existing methods have incorporated these tasks into fusion network training. These methods can be categorized into two types based on how they integrate downstream tasks: downstream task-driven fusion methods \cite{1,2,3} and downstream task-embedded fusion methods \cite{7,8,9,10}. The former directly inputs the fusion results into downstream task networks and adjusts the fusion network based on the task-specific loss. Notable examples include SeAFusion \cite{1}, TarDAL \cite{2}, and IRFS \cite{3}, with common tasks such as object detection (OD) \cite{42, 44}, semantic segmentation (SS) \cite{45, 46}, and salient object detection (SOD) \cite{41, 43}. The latter integrates the intermediate features of downstream task networks into the fusion network, refining them through task-specific adjustments or dual-task interactions to improve fusion quality. While both approaches enhance fusion quality and task performance, it remains challenging to meet the requirements of multiple downstream tasks simultaneously without retraining the fusion network (as shown in Figure \ref{fig1}(a)). Introducing multiple tasks into the training of a fusion model can increase its complexity and potentially compromise performance on specific tasks.

Faced with the above issues, one might ask: Can the output features of a fusion model be adaptively adjusted based on downstream tasks, enabling the fusion results to meet specific task requirements? If this vision is realized, retraining the fusion network for different tasks would no longer be necessary, as the output features could be dynamically modified to align with the task-specific demands. To address this challenge, as shown in Figure \ref{fig1}(b), we explore a method for constructing a fusion model that adapts to multiple downstream tasks without significantly compromising performance on any single task, offering new solutions for multi-task image fusion applications.

Specifically, we introduce downstream task-related text instructions into the feature extraction process and develop a Task-Oriented Adaptive Regulation (T-OAR) mechanism. In our network architecture, T-OAR comprises multiple Task-related Dynamic Prompt Injection (T-DPI) modules. T-DPI adaptively generates dynamic prompts based on user-input task instructions and injects these prompts into features requiring fine-tuning, guiding them to meet specific task requirements. Compared with existing methods, the proposed approach offers a significant advantage: it dynamically fine-tunes features based on task instructions without the need to retrain the network, thereby efficiently addressing the diverse needs of multiple tasks. Structurally simple yet powerful, this method demonstrates superior performance across various tasks in experimental evaluations, confirming notable improvements in adaptability, flexibility, and task specificity. Overall, this paper presents a new solution for feature extraction and fusion in multi-task environments, paving the way for IR-VIS image fusion applications in diverse scenarios. With T-OAR and embedded T-DPI modules, the constructed network adapts to different task requirements with lower computational costs and improved performance. The main contributions of this paper are summarized as follows:
\begin{itemize} 
	\item \textbf{New Requirements for IR-VIS Image Fusion.} This paper highlights the importance of generating fusion results tailored to the specific needs of downstream tasks without requiring retraining of the fusion network. The method we propose not only broadens the scope of IR-VIS image fusion research but also enhances its practical applicability, expanding the application range of fusion techniques.
	
	\item \textbf{Adaptive Adjustment Mechanism with Dynamic Prompt Injection.} We introduce a T-OAR mechanism that includes multiple T-DPI modules. This mechanism generates task-specific prompts based on user instructions and integrates them into the feature extraction process, aligning outputs precisely with task requirements and enabling flexible adaptation across diverse scenarios.
	
	\item \textbf{Flexibility Without Retraining and Efficient Structural Design.} The proposed method distinguishes itself by adapting outputs to task instructions without network retraining, allowing efficient adaptation across a variety of downstream tasks. This method is simple yet effective, demonstrating excellent performance across multiple downstream task benchmarks.

\end{itemize}
\begin{figure*}[t!]
	\centering
	\includegraphics[height=3.0in,width=6.5in]{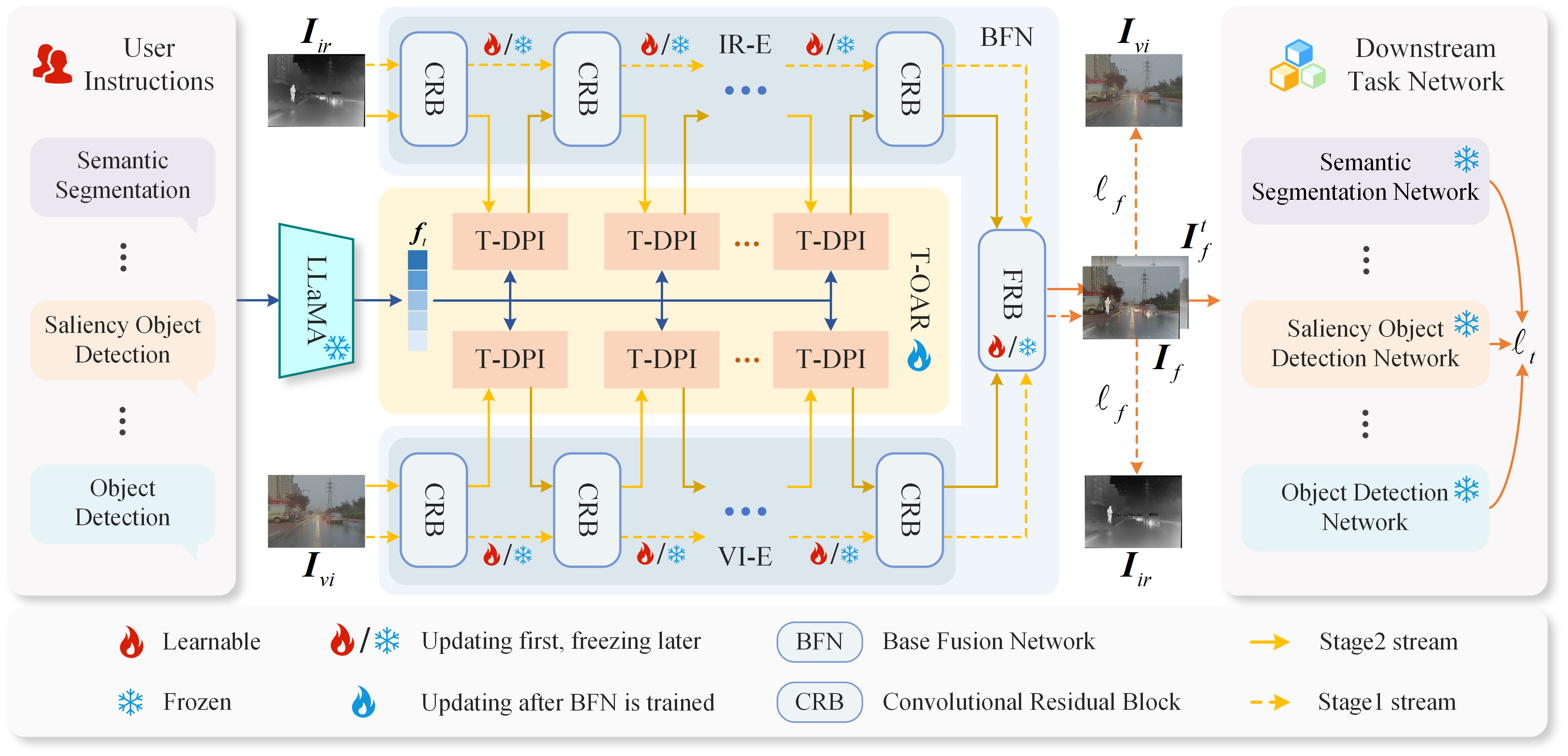}\vspace{-3mm}
	\caption{Overview of the proposed method. The proposed method integrates downstream task instruction features obtained from LLaMA into T-DPI, which generates dynamic prompts closely aligned with the specific task. This enables the method to flexibly adjust the output features of VI-E and IR-E based on input instructions, ensuring that these features meet the specific requirements of downstream tasks.}\vspace{-4.5mm}
	\label{fig2}
\end{figure*}
\section{Related Work}
\subsection{Downstream Task-Driven Fusion Methods}
Downstream task-driven fusion methods typically input the fusion results into task-specific networks and use task-specific loss functions to minimize errors, thereby indirectly optimizing the performance of the fusion network. Among these methods, SeAFusion \cite{1}, TarDAL \cite{2}, and IRFS \cite{3} optimize the fusion network by leveraging the loss functions associated with SS, OD, and SOD, respectively. However, a significant limitation of these methods is their fixed network structures, which render them inflexible to changes in downstream tasks. To address this limitation, TIMFusion \cite{4} introduces an architecture search module that allows the fusion network to dynamically adjust its structure based on different task inputs. However, TIMFusion cannot adaptively modify the trained network when the downstream task changes, requiring retraining the fusion network from scratch. This limitation restricts its deployment and application in scenarios where rapid adaptation to new tasks is crucial.

Although these methods have significantly enhanced fusion quality by integrating downstream tasks into the fusion process, they have not fully exploited the potential of these tasks to maximize fusion performance. To address this gap, BDLFusion \cite{5} introduces a Random Loss Weighting strategy, which balances the fusion loss and downstream task loss by dynamically adjusting weights. This prevents the loss most relevant to the image fusion task from dominating the optimization process. However, this method does not take into account the model's robustness to different input images, which restricts its ability to generalize. In contrast, PAIF \cite{6} generates adversarial noise based on SS results and applies it to the input source images, enhancing model generalization. Despite these advancements, these methods optimize fusion network parameters using only a single downstream task loss, making it challenging to simultaneously meet the requirements of multiple downstream tasks. This single-task focus may therefore limit the model’s flexibility and adaptability in multi-task scenarios.

\subsection{Downstream Task-Embedded Fusion Methods}
Fusion methods that embed downstream tasks typically apply the task network directly to the fused features to improve quality and achieve better results. For example, MRFS \cite{7} shares a feature fusion module between fused image reconstruction and downstream segmentation tasks to enhance fusion feature quality. DeFusion++ \cite{47} inputs the fused features into the task heads of multiple downstream task networks to support a variety of downstream tasks. However, when switching downstream tasks, the backbone network and task heads need to be fine-tuned again, which not only increases the number of training parameters but also affects deployment efficiency. Moreover, the limited interaction between fusion and task-specific features hinders further performance improvements. To address this, some methods incorporate outputs from the downstream network, such as attention maps or features, into the fusion process. DetFusion \cite{8} uses attention maps from an OD network to refine the fused features, while PSFusion \cite{9} injects SS features to enhance semantic representation. However, they does not consider whether the injected information is task-relevant. In response, SegMiF \cite{10} computes attention between semantic features from the SS network and the fused features, injecting task-specific semantic information into the fused features to better align the fusion results with downstream task requirements.

Although these methods are effective, most of them focus on ensuring that the output of the IR-VIS fusion network meets the requirements of a single downstream task, making it difficult to address the diverse needs of multiple downstream tasks simultaneously. While incorporating multiple downstream tasks into the fusion framework can help resolve this issue, it may compromise the performance of specific tasks without updating the parameters of the fusion network or downstream task networks. In contrast, this paper fully considers the unique requirements of multiple downstream tasks and, by incorporating task-specific guidance, enables a single fusion network to adapt its output features without retraining, thereby maintaining high performance across various tasks.

\vspace{-1mm}
\section{Methodology}
\label{sec:method}
\subsection{Overview}
The proposed method, illustrated in Figure \ref{fig2}, comprises two core components: the Base Fusion Network (BFN) and the T-OAR module. The BFN integrates complementary information from IR-VIS images to produce high-quality fused results. It is structured with an Infrared Feature Encoder (IR-E), a Visible Feature Encoder (VI-E), and a Fusion and Reconstruction Block (FRB). The T-OAR module adaptively adjusts the encoded features output by the BFN according to user commands for downstream tasks, ensuring that the fused images better meet the specific needs of those tasks. In terms of network architecture, the T-OAR consists of multiple T-DPI modules. Regarding model training, the method is implemented in two phases, as depicted in Figure \ref{fig2}. The dashed line represents the first training phase, which focuses on training the BFN for the initial fusion of IR-VIS images, while the solid line represents the second phase, where we freeze the pre-trained parameters of the BFN and train the T-OAR, allowing the entire fusion framework to adapt to various downstream task requirements without updating the parameters of the fusion network and downstream task network.
\subsection{Base Fusion Network}
BFN primarily consists of IR-E, VI-E, and FRB. IR-E and VI-E are used to extract features from infrared and visible images, respectively. FRB is employed to integrate information from the infrared and visible image features and reconstruct the fused image. The specific structure of BFN is illustrated in Figure \ref{fig2}. We first input $\bm{I}_{ir}$ and $\bm{I}_{vi}$ into IR-E and VI-E, respectively, and feed the obtained results into FRB. Here, IR-E and VI-E are each composed of $M$ Convolutional Residual Blocks (CRBs), and the output feature of the $i$-th CRB is denoted as $\bm{F}_{ir}^{i}$ or $\bm{F}_{vi}^{i}$. Each CRB consists of three convolutional blocks with the skip connections.

\begin{figure}[!t]
	\centering
	\includegraphics[width=0.9\linewidth]{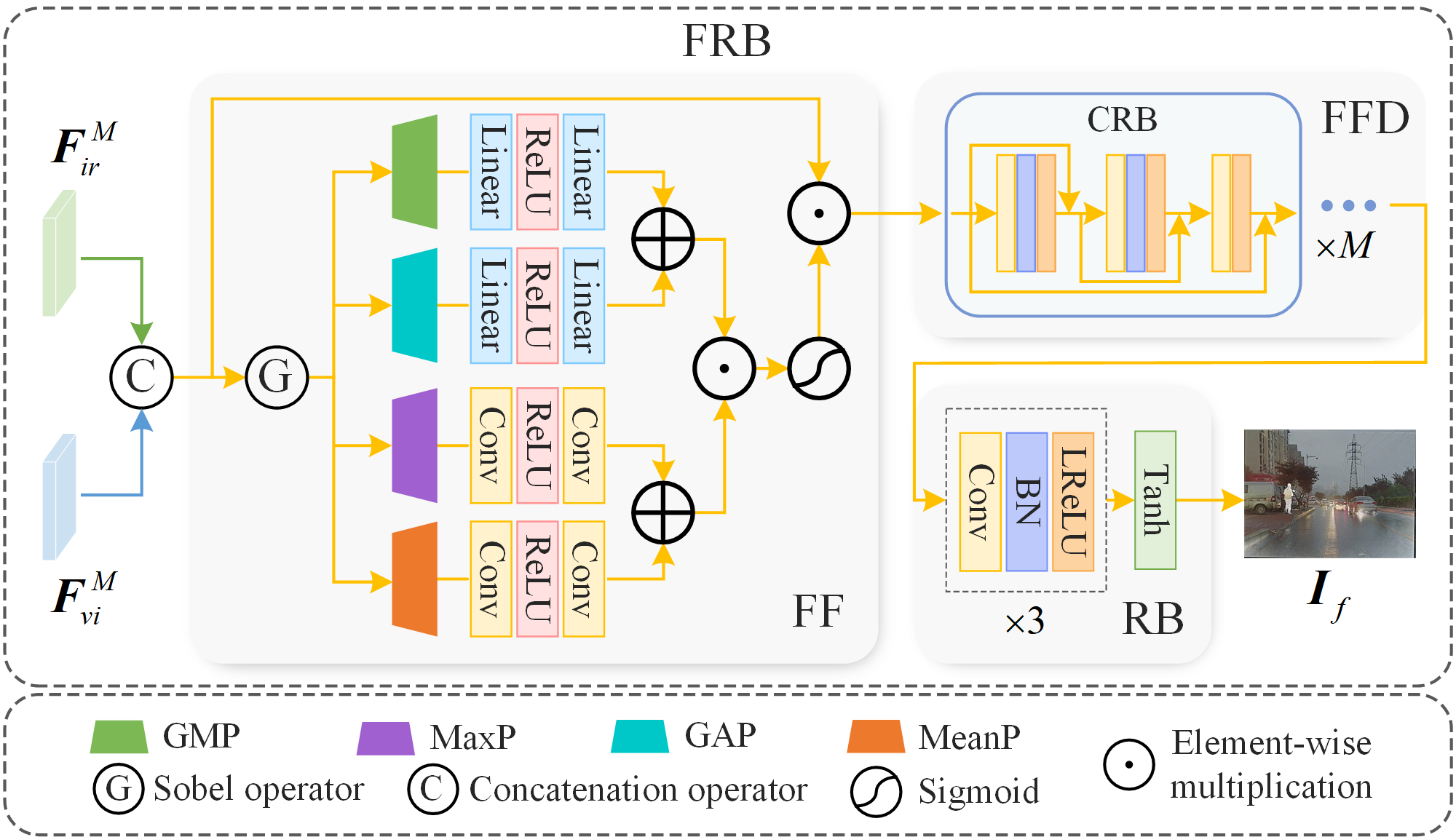}\vspace{-0.2cm}
	\caption{Structure of the FRB, including the FF, FFD, and RB submodules, where FFD consists of $M$ CRBs and RB contains 3 blocks constructed by Conv layers, BN, and LReLU. }\vspace{-0.5cm}
	\label{fig3}
\end{figure}

As illustrated in Figure \ref{fig3}, the FRB consists of a Feature Fusion (FF) block, a Fusion Feature Decoder (FFD), and a Reconstruction Block (RB). During the feature fusion process, we concatenate the infrared feature $\bm{F}_{ir}^M$ output by IR-E with the visible feature $\bm{F}_{vi}^M$ output by VI-E along the channel dimension, and then extract gradients from the concatenated features. To fully leverage the information in the gradient maps, we apply Global Max Pooling (GMP), Global Average Pooling (GAP), Max Pooling (MaxP), and Mean Pooling (MeanP) operations separately.  

Subsequently, the outputs of the GMP and GAP branches are fed into blocks composed of linear layers and ReLU, while the outputs of the MaxP and MeanP branches are processed through convolutional layers (Conv) and ReLU for feature extraction. To integrate the features obtained from multiple branches, we perform element-wise addition on the results of the GMP and GAP branches, denoted as $\bm{F}_{GMP}+\bm{F}_{GAP}$, and similarly, on the results of the MaxP and MeanP branches, denoted as $\bm{F}_{MaxP}+\bm{F}_{MeanP}$. Next, the added results are element-wise multiplied to highlight shared salient features, yielding 
\begin{equation}
     \tilde{\bm{F}} = (\bm{F}_{GMP}+\bm{F}_{GAP})\odot (\bm{F}_{MaxP}+\bm{F}_{MeanP})
\end{equation}
The results are processed through a Sigmoid function to obtain pixel-wise importance weights. Finally, these weights are element-wise multiplied with the concatenated features to obtain the fused features, which are then used by the FFD and RB to reconstruct the fused image $\bm{I}_f$:
\begin{equation}
	\bm{I}_f = \text{RB}(\text{FFD}(\text{Sigmoid}(\tilde{\bm{F}}) \odot [\bm{F}_{ir}^M, \bm{F}_{vi}^M]))
\end{equation}
where $[\cdot]$ denotes the concatenation operation.

To ensure the fused image matches the source images in brightness and gradient, we introduce a brightness loss $\ell_{bri}$ and a gradient loss $\ell_g$ to form the fusion loss $\ell_f$. Specifically, the brightness loss $\ell_{bri}$ is defined as: 
\begin{equation} 
	\ell_{bri} = \frac{1}{HW} \left\| \bm{I}_f - \max(\bm{I}_{ir}, \bm{I}_{vi}) \right\|_1 
\end{equation}
The gradient loss $\ell_g$ is defined as: 
\begin{equation} 
	\ell_g = \frac{1}{HW} \left\| \nabla \bm{I}_f - \max(\nabla \bm{I}_{ir}, \nabla \bm{I}_{vi}) \right\|_1 
\end{equation} 
Therefore, the total fusion loss is defined as: 
\begin{equation} \ell_f = \ell_g +\lambda\ell_{bri}
\end{equation} 
Here, $H$ and $W$ represent the height and width of the source images, respectively; $\lambda$ denotes the balancing parameter.

\subsection{Task-Oriented Adaptive Regulation}
T-OAR adaptively adjusts the features output by BFN according to the relevant instructions for the downstream task, ensuring that the reconstructed fusion results cater to the task's requirements. This module updates its parameters during the second stage of training. Specifically, the user-provided downstream task-related instructions (text) are input into the pre-trained large language model LLaMA \cite{24} to extract text features $\bm{f}_t$. Simultaneously, the infrared image $\bm{I}_{ir}$ and the visible image $\bm{I}_{vi}$ are fed into the first $M-1$ CRBs of IR-E and VI-E, respectively, yielding $\bm{F}_{ir}^i$ and $\bm{F}_{vi}^i$, where $i = 1, 2, \ldots, M-1$. To align the encoded features extracted by BFN with the requirements of the downstream task, both $\bm{f}_t$ and $\bm{F}_{ir/vi}^i$ are input into T-OAR for feature adjustment. T-OAR consists of $2(M-1)$ T-DPIs.

\begin{figure}[!t]
	\centering
	\includegraphics[width=0.9\linewidth]{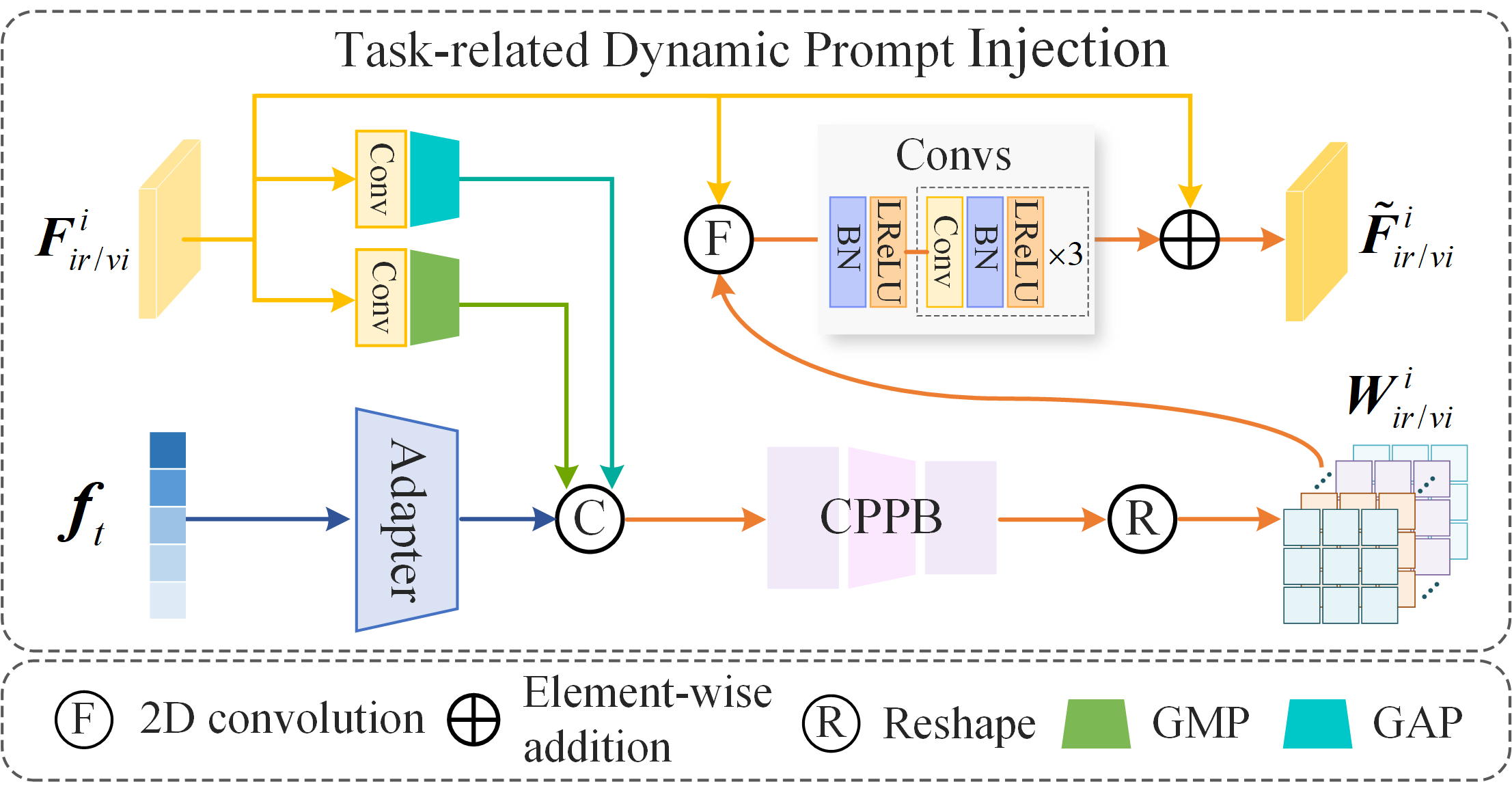}\vspace{-0.1cm}
	\caption{Structure of the T-DPI, composed of GAP, GMP, Adapter, and CPPB.}\vspace{-0.5cm}
	\label{fig4}
\end{figure}

As depicted in Figure \ref{fig4}, T-DPI primarily consists of GAP, GMP, an Adapter, and a Convolutional Parameter Prediction Block (CPPB). The Adapter, composed of two linear layers, maps $\bm{f}_t$ from the text domain to the image domain. The CPPB, consisting of two linear layers, generates dynamic convolution kernels based on the input text instructions and images. To ensure effective interaction between the text instructions and image features, we input $\bm{f}_t$ into the Adapter to obtain $\bm{f'}_t$. Concurrently, we feed ${\bm{F}}_{ir/vi}^i$ into two non-shared $3 \times 3$ Conv layers, and the resulting outputs are subjected to GAP and GMP operations, respectively, to extract the global image features ${\bm{F}}_{ir/vi}^{i,a}$ and ${\bm{F}}_{ir/vi}^{i,m}$.

To refine the semantic information from image features tailored to the needs of downstream tasks, we predict a dynamic convolution kernel that adapts to both the downstream task and the input image information. Specifically, we concatenate ${\bm{F}}_{ir/vi}^{i,a}$, ${\bm{F}}_{ir/vi}^{i,m}$, and $\bm{f'}_t$ along the channel dimension and feed the concatenated result into the CPPB. The output is then reshaped to obtain the dynamic convolution kernel ${\bm{W}}_{ir/vi}^i$: 
\begin{equation} 
	{\bm{W}}_{ir/vi}^i = \text{Reshape}(\text{CPPB}([{\bm{F}}_{ir/vi}^{i,a}, {\bm{F}}_{ir/vi}^{i,m}, {\bm{f'}_t}]))
\end{equation} 
Here, $[\cdot]$ denotes concatenation along the channel dimension. Since the parameters of the convolution kernel ${\bm{W}}_{ir/vi}^i$ can be adjusted according to the input task instructions, the proposed method enables T-OAR to adapt dynamically in response to the input instructions.

After obtaining the dynamic convolution kernel ${\bm{W}}_{ir/vi}^i$, we use it to perform a two-dimensional convolution operation on ${\bm{F}}_{ir/vi}^i$, refining the dynamic prompt ${\bm{P}}_{ir/vi}^i$ within ${\bm{F}}_{ir/vi}^i$ to suit the needs of the downstream task: 
\begin{equation} {\bm{P}}_{ir/vi}^i = \text{Convs}({\bm{W}}_{ir/vi}^i * {\bm{F}}_{ir/vi}^i) 
\end{equation} 
Here, ``$*$" denotes the two-dimensional convolution operation, and $\text{Convs}(\cdot)$ represents a convolutional block consisting of three Conv layers, four BatchNorm (BN) layers, and four LReLU layers. Since ${\bm{P}}_{ir/vi}^i$ is the dynamic prompt extracted by the dynamic convolution kernel, it contains information from both the input downstream instructions and the current features. Thus, we can use ${\bm{P}}_{ir/vi}^i$ to adjust the initial features extracted by the network, ensuring that the adjusted features meet the specific downstream task requirements, thereby enhancing downstream task performance: 
\begin{equation} 
	{\bm{\tilde{F}}}_{ir/vi}^i = {\bm{P}}_{ir/vi}^i + {\bm{F}}_{ir/vi}^i 
\end{equation} 
Here, ${\bm{\tilde{F}}}_{ir/vi}^i$ represents the features adjusted by T-DPI.

The infrared image features $\bm{F}_{ir}^i$ and visible image features $\bm{F}_{vi}^i$ at different depths are adjusted by $(M - 1)$ T-DPI modules, resulting in $\bm{\tilde{F}}_{ir}^{M - 1}$ and $\bm{\tilde{F}}_{vi}^{M - 1}$, respectively, at the $(M-1)$-th T-DPI module. We then input $\bm{\tilde{F}}_{ir}^{M - 1}$ and $\bm{\tilde{F}}_{vi}^{M-1}$ into the $M$-th CRB of IR-E and VI-E, respectively, and feed the resulting outputs into the FRB to reconstruct a fused image $\bm{I}_f^t$ tailored for downstream tasks. To enhance the performance of $\bm{I}_f^t$ in downstream tasks, we freeze the pre-trained parameters of the downstream task network corresponding to the textual instructions and input $\bm{I}_f^t$ into this network to obtain the downstream task prediction $\hat{\bm{y}}$. To train T-DPI more effectively, we introduce a task-related loss $\ell_t$ to ensure the accuracy of $\hat{\bm{y}}$:
\begin{equation} 
	\ell_t = c_t(\bm{y}_{gt}, \hat{\bm{y}}) 
\end{equation}
Here, $\bm{y}_{gt}$ represents the ground truth of $\hat{\bm{y}}$, and $c_t$ denotes the loss function used during the training of the downstream task network. For the tasks of SS, OD, and SOD, we adopt the same loss functions as those used in Segformer \cite{17}, YOLOv5 \footnote{https://github.com/ultralytics/yolov5}, and CTDNet \cite{19}, respectively.

\section{Experiments}
\subsection{Datasets}
The proposed method involves two training stages. In the first stage, we train the BFN on four public datasets, adhering to standard protocols. Specifically, we randomly select 200, 201, 217, and 230 pairs of IR-VIS images from the RoadScene \cite{22}, LLVIP \cite{23}, MSRS \cite{21}, and M$^{3}$FD \cite{2} datasets, respectively, for training. For data augmentation, we apply random flipping, rotation, and cropping to the training data. In the second stage, we train the T-OAR on the M$^{3}$FD, FMB \cite{10}, and VT5000 \cite{25} datasets to adapt the fusion network to the requirements of OD, SS, and SOD. Specifically, we use 3,150, 1,220, and 2,500 pairs of IR-VIS images from the M$^{3}$FD, FMB, and VT5000 datasets, respectively, for training. To ensure diverse training samples, we apply data augmentation techniques following the implementations of YOLOv5, Segformer \cite{17}, and CTDNet \cite{19} on these three training sets. For testing, we select 525, 280, and 2,500 pairs of IR-VIS images from the M$^{3}$FD, FMB, and VT5000 datasets, respectively, to evaluate the proposed method’s performance in OD, SS, and SOD tasks, as well as its superiority in fusion performance.

\subsection{Implementation Details}
During training, we first train the BFN, then freeze its parameters and train only the T-OAR. Both stages use the Adam optimizer \cite{28} to update the network parameters, with a batch size of 6. In the first stage, lasting 100 epochs, a dynamic learning rate adjustment strategy is employed: during the first 20 epochs, the learning rate starts at $1 \times 10^{-4}$, increases to $1 \times 10^{-3}$ over the first 20 epochs, and then decreases back to $1 \times 10^{-4}$ over the next 80 epochs. In the second stage, T-OAR is trained for 1000 epochs with a fixed learning rate of $1 \times 10^{-2}$. Additionally, the hyperparameters $M$ and $\lambda$ are set to 4 and 0.2, respectively. The experiments are implemented using the PyTorch framework and trained on a single NVIDIA GeForce RTX 4090 GPU.

\subsection{Evaluation Protocols}
We evaluate the fusion performance of the proposed method using five common objective metrics. These metrics include Cross Entropy (${Q_{CE}}$) \cite{11}, Mutual Information (${Q_{MI}}$) \cite{12}, Gradient-based Fusion Performance (${Q_{AB/F}}$) \cite{13, 38}, Chen-Blum Metric (${Q_{CB}}$) \cite{14} and Chen-Varshney Metric (${Q_{CV}}$) \cite{15}. These metrics evaluate the fusion results from different perspectives, helping to avoid potential bias associated with any single metric. Among these, lower values of ${Q_{CE}}$ and ${Q_{CV}}$ indicate higher fusion quality, while higher values of the other metrics suggest better fusion performance. Additionally, four commonly used metrics are employed to evaluate the performance of the fusion results in downstream tasks. For OD and SS tasks, we use Mean Average Precision at IoU thresholds from 0.5 to 0.95 (${\rm{mA}}{{\rm{P}}_{50 \to 95}}$) \cite{16} and Mean Pixel Intersection-over-Union (mIoU) \cite{26}, respectively. For SOD, we assess the results using the Mean F-measure (${\rm{mF}_{\beta}}$) \cite{18, 39}, Mean Absolute Error (MAE) \cite{19}, and E-measure ($\rm{E}_m$) \cite{20}.


\begin{figure*}[t!]
	\centering
	\includegraphics[height=4.0in,width=6.7in]{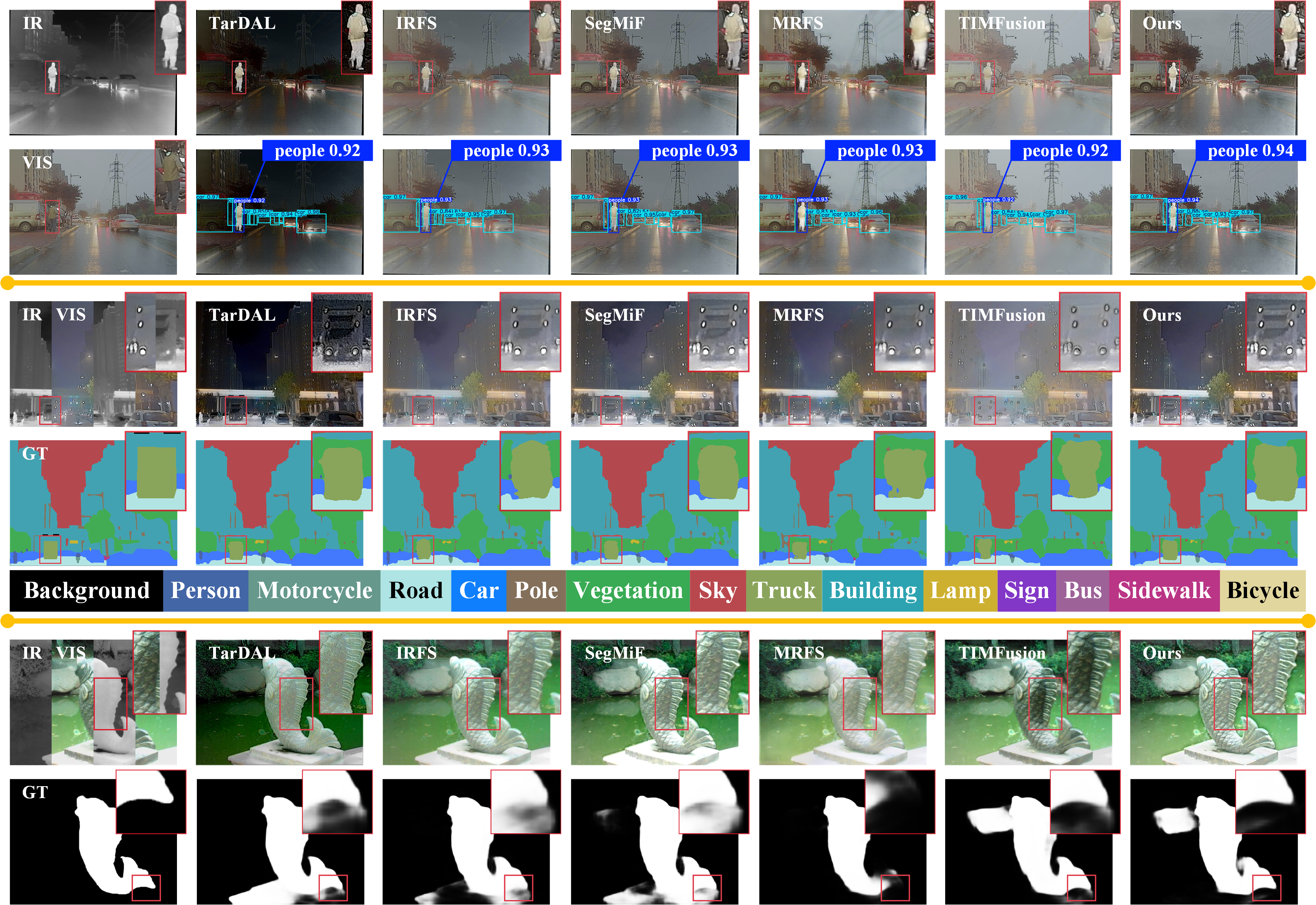}\vspace{-2mm}
	\caption{Comparison of visual effects with SOTA methods. The figure is divided into three sections, each with two rows. The input images are from the M$^{3}$FD, FMB, and VT5000 datasets, validated on OD, SS, and SOD tasks. The first column shows the IR-VIS source images and their corresponding GT for the downstream tasks. Columns two to seven display the fusion results and downstream task outcomes from the comparison methods.}\vspace{-5mm}
	\label{fig5}
\end{figure*}
\subsection{Comparison with State-of-the-art Methods}
In this study, we compare the proposed method with five state-of-the-art fusion methods (TarDAL \cite{2}, IRFS \cite{3}, SegMiF \cite{10}, MRFS \cite{7}, and TIMFusion \cite{4}) across three downstream tasks under three experimental settings. In the first setting, we evaluate non-tailored fusion results from different methods using pre-trained downstream task networks. These downstream task networks are trained with outputs from an independent fusion network, which is separate from all comparative methods. For OD, SS, and SOD tasks, we use YOLOv5s, Segformer (mit-b2), and CTDNet-18 as the downstream task networks, respectively. In the second setting, we fix the fusion network’s parameters and train separate downstream networks using its outputs, aiming to demonstrate that our method achieves comparable or superior performance without retraining. In the third setting, we compare our method with approaches jointly trained with specific downstream tasks, verifying its effectiveness across tasks and demonstrating that it closely matches the performance of joint training for each task.
\begin{table}[t!]
	\centering
	\caption{Quantitative Evaluation of Fusion Results Directly Deployed on Three Downstream Tasks. The top-ranking and second-ranking metric values are highlighted in red and blue, respectively.}\vspace{-2mm}
	\renewcommand\arraystretch{1.1}
	{\footnotesize\centerline{\tabcolsep=4.6pt
			\begin{tabular}{c|c|c|ccc}
				\hline
				\multirow{2}{*}{Methods} & OD & SS & \multicolumn{3}{c}{SOD} \\
				\cline{2-6}
				& mAP$_{50 \rightarrow 95} \uparrow$ & mIoU $\uparrow$ & mF$_{\beta} \uparrow$ & MAE $\downarrow$ & E$_m \uparrow$ \\
				\hline
				TarDAL & 0.4812 & 54.48 & 0.7767 & 0.0490 & 0.8864 \\
				IRFS & 0.5598 & 58.26 & \textcolor[rgb]{0, 0, 1}{0.8033} & 0.0417 & \textcolor[rgb]{1, 0, 0}{0.9048} \\
				SegMiF & \textcolor[rgb]{0, 0, 1}{0.6003} & \textcolor[rgb]{0, 0, 1}{60.05} & 0.8026 & \textcolor[rgb]{0, 0, 1}{0.0416} & 0.9034 \\
				MRFS & 0.5479 & 57.96 & 0.7865 & 0.0450 & 0.8939 \\
				TIMFusion & 0.5407 & 57.91 & 0.8009 & 0.0428 & 0.9017 \\
				Ours & \textcolor[rgb]{1, 0, 0}{0.6184} & \textcolor[rgb]{1, 0, 0}{60.28} & \textcolor[rgb]{1, 0, 0}{0.8058} & \textcolor[rgb]{1, 0, 0}{0.0414} & \textcolor[rgb]{0, 0, 1}{0.9043} \\
				\hline
	\end{tabular}}}\vspace{-5mm}
	\label{tab2}
\end{table}

\begin{figure}[t!]
	\centering
	\includegraphics[height=3.4in,width=3.1in]{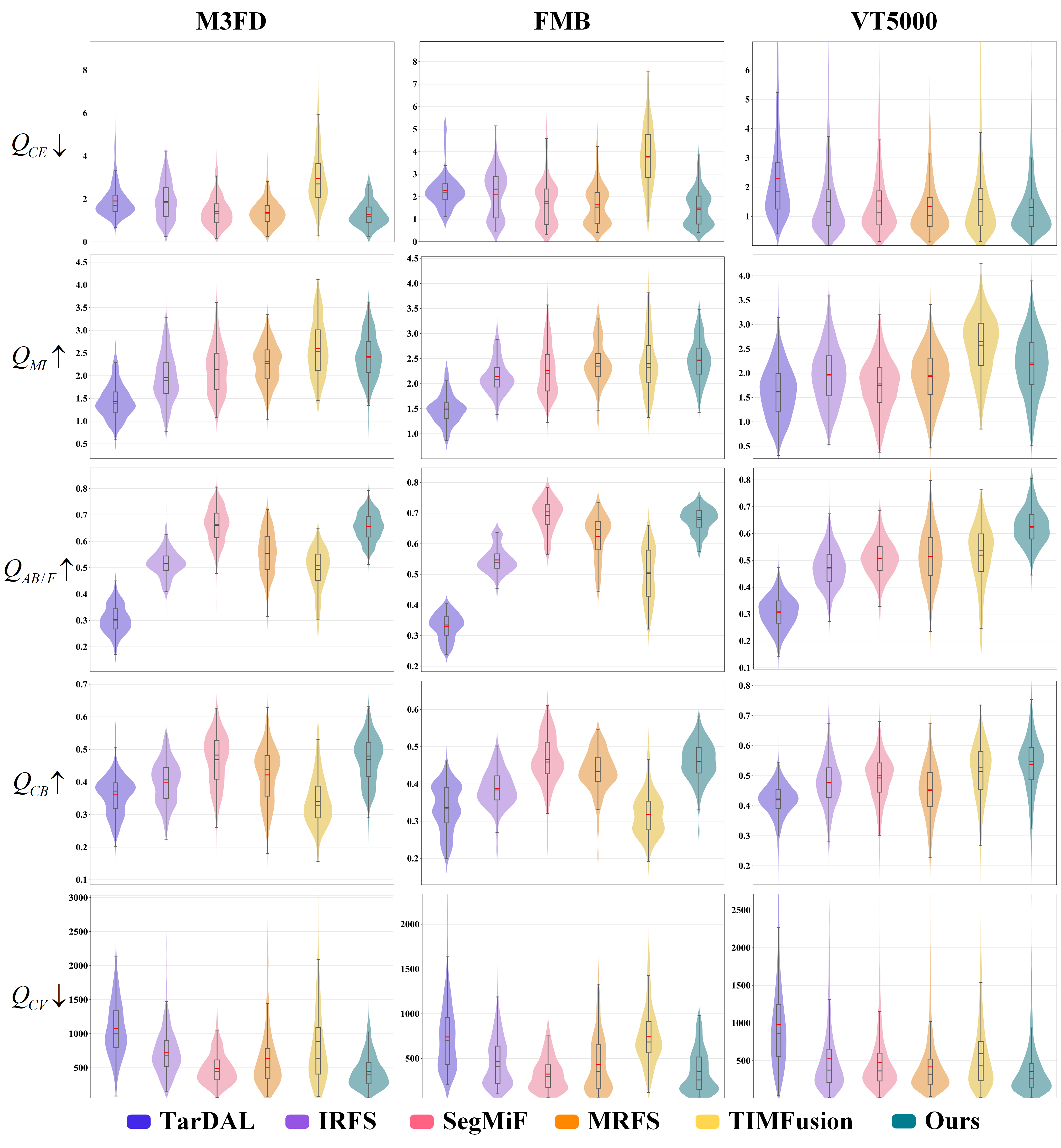}\vspace{-2mm}
	\caption{Quantitative comparison of fusion results. The x-axis represents different methods, and the y-axis shows the corresponding metric values. The upper and lower boundaries of each box denote the interquartile range, with the black line inside the box representing the median and the red line indicating the mean.}\vspace{-2mm}
	\label{fig6}
\end{figure}

\begin{figure}[t!]
	\centering
	\includegraphics[height=1.0in,width=3.0in]{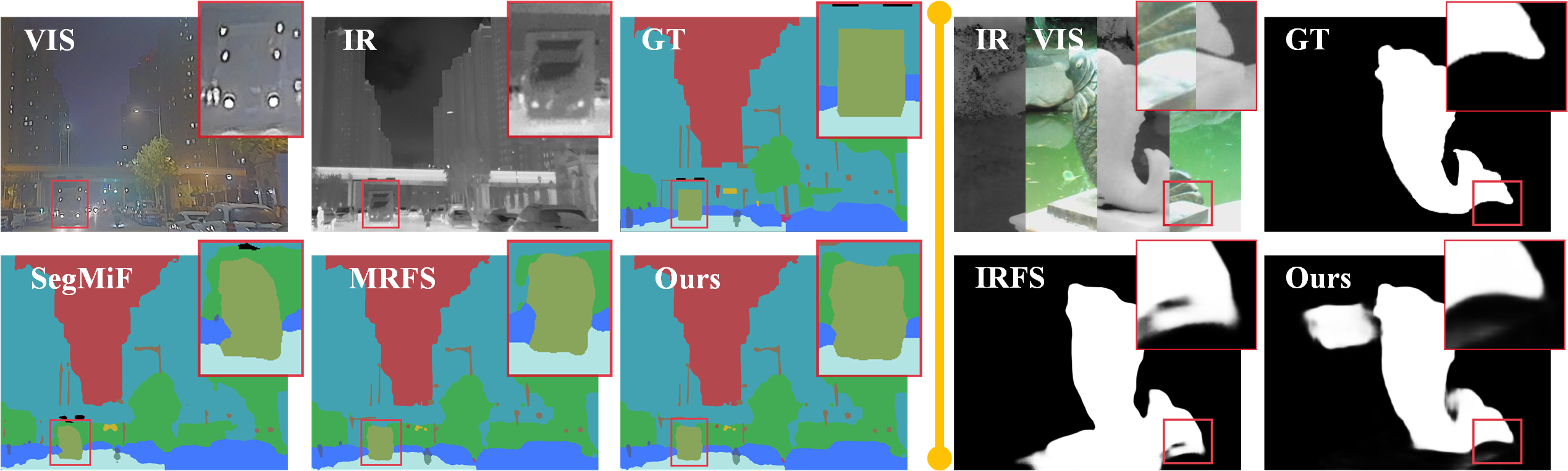}\vspace{-2mm}
	\caption{Qualitative comparison with joint training methods. The figure is divided into two parts, corresponding to the FMB and VT5000 datasets, validated on SS and SOD tasks. The first row displays the IR-VIS source images and GT, while the second row shows the downstream task results.}\vspace{-6mm}
	\label{fig7}
\end{figure}

\textbf{Experiments on Setting 1}. We directly test the fusion results obtained from TarDAL, IRFS, SegMiF, MRFS, and TIMFusion using multiple pre-trained downstream task networks to assess the impact of fusion results, which are not specifically optimized for downstream tasks, on downstream performance. As shown in Table \ref{tab2}, most of the comparison methods exhibit suboptimal performance across multiple downstream tasks. This indicates that, without specific optimization of the fusion network or downstream task networks, it is challenging for fusion results to meet the requirements of multiple downstream tasks simultaneously. In contrast, our proposed method achieves superior performance across multiple downstream tasks.

\begin{table}[htpb!]
	\centering
	\caption{Performance Evaluation of Different Methods in Setting 2. The top and second-ranking metric values are highlighted in red and blue, respectively.}\vspace{-2mm}
	\renewcommand\arraystretch{1.1}
	{\footnotesize\centerline{\tabcolsep=4.6pt
			\begin{tabular}{c|c|c|c|c|c}
				\hline
				\multirow{2}{*}{Methods} & OD    & SS    & \multicolumn{3}{c}{SOD} \\
				\cline{2-6}        &   mAP$_{50 \rightarrow 95} \uparrow$    & mIoU $\uparrow$  &   mF$_{\beta} \uparrow$   & MAE $\downarrow$   & E$_m \uparrow$ \\
				\hline
				TarDAL & 0.6157  & 59.80  & 0.7977  & 0.0426  & 0.8993  \\
				IRFS  & \textcolor[rgb]{ 1,  0,  0}{0.6306} & 59.43  & \textcolor[rgb]{ 1,  0,  0}{0.8114} & \textcolor[rgb]{ 0,  0,  1}{0.0399} & \textcolor[rgb]{ 1,  0,  0}{0.9091} \\
				SegMiF & \textcolor[rgb]{ 0,  0,  1}{0.6273} & 59.15  & 0.8023  & 0.0414  & 0.9037  \\
				MRFS  & 0.6189  & 58.28  & 0.8032  & \textcolor[rgb]{ 1,  0,  0}{0.0412} & 0.9026  \\
				TIMFusion & 0.6166  & \textcolor[rgb]{ 1,  0,  0}{60.86} & 0.7985  & 0.0428  & 0.8998  \\
				Ours  & 0.6184  & \textcolor[rgb]{ 0,  0,  1}{60.28} & \textcolor[rgb]{ 0,  0,  1}{0.8058} & 0.0414  & \textcolor[rgb]{ 0,  0,  1}{0.9043} \\
				\hline
	\end{tabular}}}\vspace{-2mm}
	\label{tab3}%
\end{table}%
\textbf{Experiments on Setting 2}. To enable downstream tasks to adapt to fusion network outputs, a common approach is to retrain the task networks on these results. To evaluate whether our method can achieve comparable or even superior performance without retraining the downstream task networks, we compare it with other retraind methods. Table \ref{tab3} presents quantitative evaluations across multiple downstream tasks. As shown, our method consistently performs well, often matching or surpassing retrained methods. For instance, in SS and SOD tasks, our method outperforms most alternatives, even though those methods are retrained for each specific task. In OD task, it closely matches the top method, with only a 0.0122 difference in ${\rm{mA}}{{\rm{P}}_{50 \to 95}}$. Notably, our method requires only 5.79M training parameters to meet the requirements of all three tasks, whereas other fusion methods require a total of 46.52M parameters due to retraining OD, SS, and SOD networks.

To further validate the fusion performance of our method across multiple downstream tasks, we visually compared the fusion results, as shown in Figure \ref {fig5}. As seen in the magnified regions, our method preserves clear edge details in target areas, such as pedestrians and vehicles, compared to other methods, greatly enhancing target distinguish ability. This clarity also brings the SS and SOD results closer to the ground truth. Additionally, to demonstrate our method’s advantages, we plot violin charts of the quantitative evaluation results, as shown in Figure \ref {fig6}. These charts reveal that our method consistently achieves higher performance across most metrics compared to all other methods. Moreover, the data points in our method’s results are more concentrated, with fewer outliers, indicating higher stability.
 
\begin{table}[htpb!]
	\centering\vspace{-2mm}
	\caption{Performance Evaluation of Different Methods in Setting 3. The top and second-ranking metric values are highlighted in red and blue, respectively.}\vspace{-2mm}
	\renewcommand\arraystretch{1.1}
	{\footnotesize\centerline{\tabcolsep=4.6pt
			\begin{tabular}{c|c|c|c|c|c}
				\hline
				\multirow{2}{*}{Methods} & OD    & SS    & \multicolumn{3}{c}{SOD} \\
				\cline{2-6}          &  mAP$_{50 \rightarrow 95} \uparrow$   & mIoU $\uparrow$  &  mF$_{\beta} \uparrow$   & MAE $\downarrow$   & E$_m \uparrow$\\
				\hline
				IRFS  & 0.5598  & 58.26  & \textcolor[rgb]{ 1,  0,  0}{0.8319} & \textcolor[rgb]{ 1,  0,  0}{0.0342} & \textcolor[rgb]{ 1,  0,  0}{0.9194} \\
				SegMiF & \textcolor[rgb]{ 0,  0,  1}{0.6003} & \textcolor[rgb]{ 1,  0,  0}{61.48} & 0.8026  & 0.0416  & 0.9034  \\
				MRFS  & 0.5479  & \textcolor[rgb]{ 0,  0,  1}{61.20} & 0.7865  & 0.0450  & 0.8939  \\
				Ours  & \textcolor[rgb]{ 1,  0,  0}{0.6184} & 60.28  & \textcolor[rgb]{ 0,  0,  1}{0.8058} & \textcolor[rgb]{ 0,  0,  1}{0.0414} & \textcolor[rgb]{ 0,  0,  1}{0.9043} \\
				\hline
	\end{tabular}}}\vspace{-2.5mm}
	\label{tab33}%
\end{table}

\textbf{Experiments on Setting 3}.
In recent years, some fusion methods, such as IRFS \cite{3}, SegMiF \cite{10}, and MRFS \cite{7}, use joint training strategies between the fusion network and downstream task networks to meet specific task requirements. However, the above methods incorporate only a single downstream task, making it difficult to meet the needs of multiple tasks. We conduct both qualitative and quantitative comparisons with these methods. As shown in Table \ref{tab33}, while these joint training methods perform well on their respective trained tasks, their performance in other tasks is relatively poor. For example, although SegMiF achieves the highest mIoU in SS (the task included in its training), its performance in OD and SOD (the tasks not included in the training) is notably lower than that of our method. Similar patterns appear with IRFS and MRFS, with the former trained for SOD and the latter for SS.

By contrast, our method consistently performs well across multiple downstream tasks and requires significantly fewer training parameters than joint training methods, making it better suited for deployment on platforms with limited computational resources. To further demonstrate the qualitative advantages of our method, we visualize the downstream task results. As shown in Figure \ref {fig7}, our method produces SS results that align closely with the ground truth, with fewer misclassified pixels. In the last row of Figure \ref{fig7}, the zoomed-in areas of SOD show that our method accurately detects stone sculpture. In summary, with fewer training parameters, our method clearly outperforms joint training methods in addressing multiple downstream tasks. \textbf{It is worth noting that additional experimental results, including visual demonstrations of the fused images, objective evaluation data, and parameter analysis, can be found in the supplementary materials.}

\subsection{Ablation Study}
The proposed method includes two main parts: FRB and T-OAR. FRB's core component is FF, while T-DPI is central to T-OAR. To verify their effectiveness, we perform ablation studies on the VT5000 dataset, assessing fusion results' visual quality and their impact on downstream tasks. Due to space constraints, \textbf{visual quality results are provided in the supplementary materials}.
\begin{table}[htbp!]
	\centering\vspace{-2mm}
	\caption{Performance of different models on downstream tasks. The evaluation metric value ranked first is highlighted in red.}\vspace{-2mm}
	\renewcommand\arraystretch{1.1}
	{\footnotesize\centerline{\tabcolsep=4.6pt
			\begin{tabular}{c|c|c|ccc}
				\hline
				\multirow{2}{*}{Models} & OD    & SS    & \multicolumn{3}{c}{SOD} \\
				\cline{2-6}          & mAP$_{50 \rightarrow 95} \uparrow$   & mIoU $\uparrow$  & mF$_{\beta} \uparrow$    & MAE $\downarrow$   & E$_m \uparrow$ \\
				\hline
				I     & 0.6147  & 59.80  & 0.8051  & 0.0421  & 0.9031  \\
				II    & 0.6134  & 58.67  & 0.8010  & 0.0438  & 0.9011  \\
				III   & 0.5868  & 55.82  & 0.8036  & 0.0419  & 0.9041  \\
				IV    & 0.6127  & 59.18  & 0.7937  & 0.0459  & 0.8953  \\
				V     & 0.6103  & 59.08  & 0.7960  & 0.0451  & 0.8979  \\
				VI    & 0.6181  & 59.86  & 0.8003  & 0.0440  & 0.8996  \\
				Ours  & \textcolor[rgb]{ 1,  0,  0}{0.6184} & \textcolor[rgb]{ 1,  0,  0}{60.28} & \textcolor[rgb]{ 1,  0,  0}{0.8058} & \textcolor[rgb]{ 1,  0,  0}{0.0414} & \textcolor[rgb]{ 1,  0,  0}{0.9043} \\
				\hline
	\end{tabular}}}\vspace{-5.5mm}
	\label{tab4}%
\end{table}%

\textbf{Effectiveness of FF}. FF enhances and fuses features through gradient maps, resulting in fusion outputs with richer texture details. To verify the effectiveness of FF, we replace it with a concatenation operation along the channel dimension, creating a model named Model I. As shown in Table \ref{tab4}, Model I demonstrates decreased performance across three downstream tasks compared to the complete model, confirming the effectiveness of FF.

\textbf{Effectiveness Task Instructions}. T-DPI refines BFN features based on downstream task instructions, enabling fusion results to adapt to various tasks. To test the effectiveness of these instructions, we remove them and used CNNs to refine the BFN features, creating Model II. As shown in Table \ref{tab4}, Model II performs worse than the complete model, confirming the importance of task instructions.

\textbf{Effectiveness of T-DPI}. To further validate the effectiveness of T-DPI, we create three comparison models: Model III, where T-DPI and task instructions are removed, retaining only the BFN; Model IV, where T-DPI is removed, and text features are added to the BFN output and processed with CNNs; and Model V, where T-DPI is removed, and text features are concatenated with BFN features and processed with CNNs. Table \ref{tab4} shows that Model III performs poorly on multiple tasks, whereas the complete model performs better across all downstream tasks. Although Models IV and V perform well on OD task, their performance on the other tasks is weaker.

In T-DPI, GAP and GMP aggregate global image information to refine BFN output based on task instructions. To assess their effectiveness, we design Model VI, which removes GAP and GMP, relying only on task instructions. As shown in Table \ref{tab4}, Model VI underperforms across tasks compared to the complete model.

\section{Conclusion}
This paper proposes a novel method for adaptive image fusion in multi-task environments through the introduction of T-OAR. By incorporating task-specific instructions into the feature extraction process, our method allows for dynamic fine-tuning of fusion outputs without the need for retraining the network. This not only enhances the model's adaptability and flexibility across different downstream tasks but also reduces computational costs while maintaining high performance. The experimental results demonstrate the effectiveness of our method in meeting the diverse requirements of multiple tasks, making it a promising solution for image fusion applications in real-world scenarios.
{
    \small
    \bibliographystyle{ieeenat_fullname}
    \bibliography{main}

\begin{thebibliography}{46}
\providecommand{\natexlab}[1]{#1}
\providecommand{\url}[1]{\texttt{#1}}
\expandafter\ifx\csname urlstyle\endcsname\relax
  \providecommand{\doi}[1]{doi: #1}\else
  \providecommand{\doi}{doi: \begingroup \urlstyle{rm}\Url}\fi

\bibitem[Arbeláez et~al.(2011)Arbeláez, Maire, Fowlkes, and Malik]{18}
Pablo Arbeláez, Michael Maire, Charless Fowlkes, and Jitendra Malik.
\newblock Contour detection and hierarchical image segmentation.
\newblock \emph{IEEE Transactions on Pattern Analysis and Machine
  Intelligence}, 33\penalty0 (5):\penalty0 898--916, 2011.

\bibitem[Borji(2015)]{41}
Ali Borji.
\newblock What is a salient object? a dataset and a baseline model for salient
  object detection.
\newblock \emph{IEEE Transactions on Image Processing}, 24\penalty0
  (2):\penalty0 742--756, 2015.

\bibitem[Bulanon et~al.(2009)Bulanon, Burks, and Alchanatis]{11}
D.M. Bulanon, T.F. Burks, and V. Alchanatis.
\newblock Image fusion of visible and thermal images for fruit detection.
\newblock \emph{Biosystems Engineering}, 103\penalty0 (1):\penalty0 12--22,
  2009.

\bibitem[Chen and Varshney(2007)]{15}
Hao Chen and Pramod~K. Varshney.
\newblock A human perception inspired quality metric for image fusion based on
  regional information.
\newblock \emph{Information Fusion}, 8\penalty0 (2):\penalty0 193--207, 2007.

\bibitem[Chen and Blum(2009)]{14}
Yin Chen and Rick~S. Blum.
\newblock A new automated quality assessment algorithm for image fusion.
\newblock \emph{Image and Vision Computing}, 27\penalty0 (10):\penalty0
  1421--1432, 2009.

\bibitem[Deng-Ping~Fan(2018)]{20}
Yang Cao Bo Ren Ming-Ming Cheng Ali~Borji Deng-Ping~Fan, Cheng~Gong.
\newblock Enhanced-alignment measure for binary foreground map evaluation.
\newblock In \emph{International Joint Conferences on Artificial Intelligence},
  2018.

\bibitem[Fu et~al.(2024)Fu, Wang, Duan, Xiao, Dian, Li, and Li]{42}
Haolong Fu, Shixun Wang, Puhong Duan, Changyan Xiao, Renwei Dian, Shutao Li,
  and Zhiyong Li.
\newblock Lraf-net: Long-range attention fusion network for visible–infrared
  object detection.
\newblock \emph{IEEE Transactions on Neural Networks and Learning Systems},
  35\penalty0 (10):\penalty0 13232--13245, 2024.

\bibitem[He and Todorovic(2022)]{16}
Liqiang He and Sinisa Todorovic.
\newblock Destr: Object detection with split transformer.
\newblock In \emph{2022 IEEE/CVF Conference on Computer Vision and Pattern
  Recognition (CVPR)}, pages 9367--9376, 2022.

\bibitem[Jia et~al.(2021)Jia, Zhu, Li, Tang, and Zhou]{23}
Xinyu Jia, Chuang Zhu, Minzhen Li, Wenqi Tang, and Wenli Zhou.
\newblock Llvip: A visible-infrared paired dataset for low-light vision.
\newblock In \emph{Proceedings of the IEEE/CVF International Conference on
  Computer Vision Workshops (ICCVW)}, pages 3496--3504, 2021.

\bibitem[Kingma and Ba(2015)]{28}
Diederik~P. Kingma and Jimmy Ba.
\newblock Adam: A method for stochastic optimization.
\newblock In \emph{International Conference on Learning Representations
  (ICLR)}, 2015.

\bibitem[Li et~al.(2020)Li, Wu, and Kittler]{40}
Hui Li, Xiao-Jun Wu, and Josef Kittler.
\newblock Mdlatlrr: A novel decomposition method for infrared and visible image
  fusion.
\newblock \emph{IEEE Transactions on Image Processing}, 29:\penalty0
  4733--4746, 2020.

\bibitem[Li et~al.(2021{\natexlab{a}})Li, Cen, Liu, Chen, and Yu]{32}
Huafeng Li, Yueliang Cen, Yu Liu, Xun Chen, and Zhengtao Yu.
\newblock Different input resolutions and arbitrary output resolution: A meta
  learning-based deep framework for infrared and visible image fusion.
\newblock \emph{IEEE Transactions on Image Processing}, 30:\penalty0
  4070--4083, 2021{\natexlab{a}}.

\bibitem[Li et~al.(2023{\natexlab{a}})Li, Xu, Wu, Lu, and Kittler]{37}
Hui Li, Tianyang Xu, Xiao-Jun Wu, Jiwen Lu, and Josef Kittler.
\newblock Lrrnet: A novel representation learning guided fusion network for
  infrared and visible images.
\newblock \emph{IEEE Transactions on Pattern Analysis and Machine
  Intelligence}, 45\penalty0 (9):\penalty0 11040--11052, 2023{\natexlab{a}}.

\bibitem[Li et~al.(2023{\natexlab{b}})Li, Zhao, Li, Yu, and Lu]{31}
Huafeng Li, Junzhi Zhao, Jinxing Li, Zhengtao Yu, and Guangming Lu.
\newblock Feature dynamic alignment and refinement for infrared–visible image
  fusion: Translation robust fusion.
\newblock \emph{Information Fusion}, 95:\penalty0 26--41, 2023{\natexlab{b}}.

\bibitem[Li et~al.(2024{\natexlab{a}})Li, Liu, Zhang, and Liu]{30}
Huafeng Li, Junyu Liu, Yafei Zhang, and Yu Liu.
\newblock A deep learning framework for infrared and visible image fusion
  without strict registration.
\newblock \emph{International Journal of Computer Vision}, 132:\penalty0
  1625–1644, 2024{\natexlab{a}}.

\bibitem[Li et~al.(2024{\natexlab{b}})Li, Liu, Li, and Tan]{36}
Xilai Li, Wuyang Liu, Xiaosong Li, and Haishu Tan.
\newblock Physical perception network and an all-weather multi-modality
  benchmark for adverse weather image fusion.
\newblock \emph{arXiv preprint arXiv: 2402. 02090}, 2024{\natexlab{b}}.

\bibitem[Li et~al.(2021{\natexlab{b}})Li, Pang, Cao, Shen, and Shao]{44}
Yazhao Li, Yanwei Pang, Jiale Cao, Jianbing Shen, and Ling Shao.
\newblock Improving single shot object detection with feature scale unmixing.
\newblock \emph{IEEE Transactions on Image Processing}, 30:\penalty0
  2708--2721, 2021{\natexlab{b}}.

\bibitem[Liang et~al.(2024)Liang, Jiang, Ma, Liu, and Ma]{47}
Pengwei Liang, Junjun Jiang, Qing Ma, Xianming Liu, and Jiayi Ma.
\newblock Fusion from decomposition: A self-supervised approach for image
  fusion and beyond.
\newblock \emph{arXiv preprint arXiv: 2410.12274}, 2024.

\bibitem[Liu et~al.(2022)Liu, Fan, Huang, Wu, Liu, Zhong, and Luo]{2}
Jinyuan Liu, Xin Fan, Zhanbo Huang, Guanyao Wu, Risheng Liu, Wei Zhong, and
  Zhongxuan Luo.
\newblock Target-aware dual adversarial learning and a multi-scenario
  multi-modality benchmark to fuse infrared and visible for object detection.
\newblock In \emph{Proceedings of the IEEE/CVF Conference on Computer Vision
  and Pattern Recognition (CVPR)}, pages 5802--5811, 2022.

\bibitem[Liu et~al.(2023{\natexlab{a}})Liu, Liu, Wu, Ma, Liu, Zhong, Luo, and
  Fan]{10}
Jinyuan Liu, Zhu Liu, Guanyao Wu, Long Ma, Risheng Liu, Wei Zhong, Zhongxuan
  Luo, and Xin Fan.
\newblock Multi-interactive feature learning and a full-time multi-modality
  benchmark for image fusion and segmentation.
\newblock In \emph{2023 IEEE/CVF International Conference on Computer Vision
  (ICCV)}, pages 8081--8090, 2023{\natexlab{a}}.

\bibitem[Liu et~al.(2024{\natexlab{a}})Liu, Lin, Wu, Liu, Luo, and Fan]{33}
Jinyuan Liu, Runjia Lin, Guanyao Wu, Risheng Liu, Zhongxuan Luo, and Xin Fan.
\newblock Coconet: Coupled contrastive learning network with multi-level
  feature ensemble for multi-modality image fusion.
\newblock \emph{International Journal of Computer Vision}, 132\penalty0
  (5):\penalty0 1748--1775, 2024{\natexlab{a}}.

\bibitem[Liu et~al.(2024{\natexlab{b}})Liu, Liu, Liu, Fan, and Luo]{4}
Risheng Liu, Zhu Liu, Jinyuan Liu, Xin Fan, and Zhongxuan Luo.
\newblock A task-guided, implicitly-searched and meta-initialized deep model
  for image fusion.
\newblock \emph{IEEE Transactions on Pattern Analysis and Machine
  Intelligence}, 46\penalty0 (10):\penalty0 6594--6609, 2024{\natexlab{b}}.

\bibitem[Liu et~al.(2011)Liu, Yuan, Sun, Wang, Zheng, Tang, and Shum]{39}
Tie Liu, Zejian Yuan, Jian Sun, Jingdong Wang, Nanning Zheng, Xiaoou Tang, and
  Heung-Yeung Shum.
\newblock Learning to detect a salient object.
\newblock \emph{IEEE Transactions on Pattern Analysis and Machine
  Intelligence}, 33\penalty0 (2):\penalty0 353--367, 2011.

\bibitem[Liu et~al.(2024{\natexlab{c}})Liu, Zeng, Tao, and Fang]{45}
Yue Liu, Jun Zeng, Xingzhen Tao, and Gang Fang.
\newblock Rethinking self-supervised semantic segmentation: Achieving
  end-to-end segmentation.
\newblock \emph{IEEE Transactions on Pattern Analysis and Machine
  Intelligence}, 46\penalty0 (12):\penalty0 10036--10046, 2024{\natexlab{c}}.

\bibitem[Liu et~al.(2023{\natexlab{b}})Liu, Liu, Wu, Ma, Fan, and Liu]{5}
Zhu Liu, Jinyuan Liu, Guanyao Wu, Long Ma, Xin Fan, and Risheng Liu.
\newblock Bi-level dynamic learning for jointly multi-modality image fusion and
  beyond.
\newblock In \emph{Proceedings of the Thirty-Second International Joint
  Conference on Artificial Intelligence, {IJCAI-23}}, pages 1240--1248,
  2023{\natexlab{b}}.

\bibitem[Liu et~al.(2023{\natexlab{c}})Liu, Liu, Zhang, Ma, Fan, and Liu]{6}
Zhu Liu, Jinyuan Liu, Benzhuang Zhang, Long Ma, Xin Fan, and Risheng Liu.
\newblock Paif: Perception-aware infrared-visible image fusion for
  attack-tolerant semantic segmentation.
\newblock In \emph{Proceedings of the 31st ACM International Conference on
  Multimedia}, page 3706–3714, 2023{\natexlab{c}}.

\bibitem[Ma et~al.(2021)Ma, Li, Chen, Hao, and Qin]{43}
Guangxiao Ma, Shuai Li, Chenglizhao Chen, Aimin Hao, and Hong Qin.
\newblock Rethinking image salient object detection: Object-level semantic
  saliency reranking first, pixelwise saliency refinement later.
\newblock \emph{IEEE Transactions on Image Processing}, 30:\penalty0
  4238--4252, 2021.

\bibitem[Ma et~al.(2019)Ma, Ma, and Li]{38}
Jiayi Ma, Yong Ma, and Chang Li.
\newblock Infrared and visible image fusion methods and applications: A survey.
\newblock \emph{Information Fusion}, 45:\penalty0 153--178, 2019.

\bibitem[Sun et~al.(2022)Sun, Cao, Zhu, and Hu]{8}
Yiming Sun, Bing Cao, Pengfei Zhu, and Qinghua Hu.
\newblock Detfusion: A detection-driven infrared and visible image fusion
  network.
\newblock In \emph{Proceedings of the 30th ACM International Conference on
  Multimedia}, page 4003–4011, 2022.

\bibitem[Tang et~al.(2022{\natexlab{a}})Tang, Yuan, and Ma]{1}
Linfeng Tang, Jiteng Yuan, and Jiayi Ma.
\newblock Image fusion in the loop of high-level vision tasks: A semantic-aware
  real-time infrared and visible image fusion network.
\newblock \emph{Information Fusion}, 82:\penalty0 28--42, 2022{\natexlab{a}}.

\bibitem[Tang et~al.(2022{\natexlab{b}})Tang, Yuan, Zhang, Jiang, and Ma]{21}
Linfeng Tang, Jiteng Yuan, Hao Zhang, Xingyu Jiang, and Jiayi Ma.
\newblock Piafusion: A progressive infrared and visible image fusion network
  based on illumination aware.
\newblock \emph{Information Fusion}, 83-84:\penalty0 79--92,
  2022{\natexlab{b}}.

\bibitem[Tang et~al.(2023)Tang, Zhang, Xu, and Ma]{9}
Linfeng Tang, Hao Zhang, Han Xu, and Jiayi Ma.
\newblock Rethinking the necessity of image fusion in high-level vision tasks:
  A practical infrared and visible image fusion network based on progressive
  semantic injection and scene fidelity.
\newblock \emph{Information Fusion}, 99:\penalty0 101870, 2023.

\bibitem[Touvron et~al.(2023)Touvron, Lavril, Izacard, Martinet, Lachaux,
  Lacroix, Rozière, Goyal, Hambro, Azhar, Rodriguez, Joulin, Grave, and
  Lample]{24}
Hugo Touvron, Thibaut Lavril, Gautier Izacard, Xavier Martinet, Marie-Anne
  Lachaux, Timothée Lacroix, Baptiste Rozière, Naman Goyal, Eric Hambro,
  Faisal Azhar, Aurelien Rodriguez, Armand Joulin, Edouard Grave, and Guillaume
  Lample.
\newblock Llama: Open and efficient foundation language models.
\newblock \emph{arXiv preprint arXiv: 2302.13971}, 2023.

\bibitem[Tu et~al.(2023)Tu, Ma, Li, Li, Xu, and Liu]{25}
Zhengzheng Tu, Yan Ma, Zhun Li, Chenglong Li, Jieming Xu, and Yongtao Liu.
\newblock Rgbt salient object detection: A large-scale dataset and benchmark.
\newblock \emph{IEEE Transactions on Multimedia}, 25:\penalty0 4163--4176,
  2023.

\bibitem[Wang et~al.(2023)Wang, Liu, Liu, and Fan]{3}
Di Wang, Jinyuan Liu, Risheng Liu, and Xin Fan.
\newblock An interactively reinforced paradigm for joint infrared-visible image
  fusion and saliency object detection.
\newblock \emph{Information Fusion}, 98:\penalty0 101828, 2023.

\bibitem[Wu et~al.(2023)Wu, Fang, He, He, Ma, and Zhong]{46}
Linshan Wu, Leyuan Fang, Xingxin He, Min He, Jiayi Ma, and Zhun Zhong.
\newblock Querying labeled for unlabeled: Cross-image semantic consistency
  guided semi-supervised semantic segmentation.
\newblock \emph{IEEE Transactions on Pattern Analysis and Machine
  Intelligence}, 45\penalty0 (7):\penalty0 8827--8844, 2023.

\bibitem[Xie et~al.(2021)Xie, Wang, Yu, Anandkumar, Alvarez, and Luo]{17}
Enze Xie, Wenhai Wang, Zhiding Yu, Anima Anandkumar, Jose~M Alvarez, and Ping
  Luo.
\newblock Segformer: Simple and efficient design for semantic segmentation with
  transformers.
\newblock In \emph{Neural Information Processing Systems (NeurIPS)}, 2021.

\bibitem[Xu et~al.(2022)Xu, Ma, Jiang, Guo, and Ling]{22}
Han Xu, Jiayi Ma, Junjun Jiang, Xiaojie Guo, and Haibin Ling.
\newblock U2fusion: A unified unsupervised image fusion network.
\newblock \emph{IEEE Transactions on Pattern Analysis and Machine
  Intelligence}, 44\penalty0 (1):\penalty0 502--518, 2022.

\bibitem[Xydeas et~al.(2000)Xydeas, Petrovic, et~al.]{13}
Costas~S Xydeas, Vladimir Petrovic, et~al.
\newblock Objective image fusion performance measure.
\newblock \emph{Electronics letters}, 36\penalty0 (4):\penalty0 308--309, 2000.

\bibitem[Yang et~al.(2021)Yang, Liu, Huang, Wan, Wen, and Guan]{34}
Yong Yang, Jiaxiang Liu, Shuying Huang, Weiguo Wan, Wenying Wen, and Juwei
  Guan.
\newblock Infrared and visible image fusion via texture conditional generative
  adversarial network.
\newblock \emph{IEEE Transactions on Circuits and Systems for Video
  Technology}, 31\penalty0 (12):\penalty0 4771--4783, 2021.

\bibitem[Yu et~al.(2021)Yu, Gao, Wang, Yu, Shen, and Sang]{26}
Changqian Yu, Changxin Gao, Jingbo Wang, Gang Yu, Chunhua Shen, and Nong Sang.
\newblock Bisenet v2: Bilateral network with guided aggregation for real-time
  semantic segmentation.
\newblock \emph{International journal of computer vision}, 129:\penalty0
  3051--3068, 2021.

\bibitem[Zamir et~al.(2022)Zamir, Arora, Khan, Hayat, Khan, and Yang]{29}
Syed~Waqas Zamir, Aditya Arora, Salman Khan, Munawar Hayat, Fahad~Shahbaz Khan,
  and Ming–Hsuan Yang.
\newblock Restormer: Efficient transformer for high-resolution image
  restoration.
\newblock In \emph{2022 IEEE/CVF Conference on Computer Vision and Pattern
  Recognition (CVPR)}, pages 5718--5729, 2022.

\bibitem[Zhang et~al.(2024)Zhang, Zuo, Jiang, Guo, and Ma]{7}
Hao Zhang, Xuhui Zuo, Jie Jiang, Chunchao Guo, and Jiayi Ma.
\newblock Mrfs: Mutually reinforcing image fusion and segmentation.
\newblock In \emph{2024 IEEE/CVF Conference on Computer Vision and Pattern
  Recognition (CVPR)}, pages 26964--26973, 2024.

\bibitem[Zhang et~al.(2020)Zhang, Ye, and Xiao]{12}
Xingchen Zhang, Ping Ye, and Gang Xiao.
\newblock Vifb: A visible and infrared image fusion benchmark.
\newblock In \emph{2020 IEEE/CVF Conference on Computer Vision and Pattern
  Recognition Workshops (CVPRW)}, pages 468--478, 2020.

\bibitem[Zhao et~al.(2023)Zhao, Xie, Zhao, He, and Lu]{35}
Wenda Zhao, Shigeng Xie, Fan Zhao, You He, and Huchuan Lu.
\newblock Metafusion: Infrared and visible image fusion via meta-feature
  embedding from object detection.
\newblock In \emph{Proceedings of the IEEE/CVF Conference on Computer Vision
  and Pattern Recognition (CVPR)}, pages 13955--13965, 2023.

\bibitem[Zhao et~al.(2021)Zhao, Xia, Xie, and Li]{19}
Zhirui Zhao, Changqun Xia, Chenxi Xie, and Jia Li.
\newblock Complementary trilateral decoder for fast and accurate salient object
  detection.
\newblock In \emph{Proceedings of the 29th ACM International Conference on
  Multimedia}, page 4967–4975, 2021.

\end{thebibliography}
}


\end{document}